\documentclass[journal]{IEEEtran}
\ifCLASSINFOpdf
\else
\fi
\usepackage{graphicx}
\usepackage{hyperref}
\usepackage{amssymb}
\usepackage{amsmath}
\usepackage{lineno}
\usepackage{array}
\usepackage{caption}
\usepackage{subcaption}
\usepackage{multirow}
\usepackage{csquotes}
\usepackage{amsfonts}

\begin{document}

\title{Monitoring COVID-19 social distancing with person detection and tracking via fine-tuned YOLO v3 and Deepsort techniques}
%
%
%

\author{Narinder Singh Punn, Sanjay Kumar Sonbhadra, Sonali Agarwal and Gaurav Rai
\thanks{N. S. Punn, S. K. Sonbhadra, S. Agarwal, Indian Institute of Information Technology Allahabad, Jhalwa, Prayagraj, Uttar Pradesh, India; emails: \{pse2017002, rsi2017502, sonali\}@iiita.ac.in.}
\thanks{G. Rai, Data Analytics, Comptroller and Auditor General of India; email: raig@cag.gov.in.}
}

\maketitle

\begin{abstract}
The rampant coronavirus disease 2019 (COVID-19) has brought global crisis with its deadly spread to more than 180 countries, and about 3,519,901 confirmed cases along with 247,630 deaths globally as on May 4, 2020. The absence of any active therapeutic agents and the lack of immunity against COVID-19 increases the vulnerability of the population. Since there are no vaccines available, social distancing is the only feasible approach to fight against this pandemic. Motivated by this notion, this article proposes a deep learning based framework for automating the task of monitoring social distancing using surveillance video. The proposed framework utilizes the YOLO v3 object detection model to segregate humans from the background and Deepsort approach to track the identified people with the help of bounding boxes and assigned IDs. The results of the YOLO v3 model are further compared with other popular state-of-the-art models, e.g. faster region-based CNN (convolution neural network) and single shot detector (SSD) in terms of mean average precision (mAP), frames per second (FPS) and loss values defined by object classification and localization. Later, the pairwise vectorized \textit{L2} norm is computed based on the three-dimensional feature space obtained by using the centroid coordinates and dimensions of the bounding box. The violation index term is proposed to quantize the non adoption of social distancing protocol. From the experimental analysis, it is observed that the YOLO v3 with Deepsort tracking scheme displayed best results with balanced mAP and FPS score to monitor the social distancing in real-time. \end{abstract}
\begin{IEEEkeywords}
COVID-19, Video surveillance, Social distancing, Object detection, Object tracking.
\end{IEEEkeywords}
\section{Introduction}
\IEEEPARstart{C}{OVID-19} belongs to the family of coronavirus caused diseases, initially reported at Wuhan, China, during late December 2020. On March 11, it spread over 114 countries with 118,000 active cases and 4000 deaths, WHO declared this a pandemic~\cite{bworld, world2020director}. On May 4, 2020, over 3,519,901 cases and 247,630 deaths had been reported worldwide. Several healthcare organizations, medical experts and scientists are trying to develop proper medicines and vaccines for this deadly virus, but till date, no success is reported. This situation forces the global community to look for alternate ways to stop the spread of this infectious virus. Social distancing is claimed as the best spread stopper in the present scenario, and all affected countries are locked-down to implement social distancing. This research is aimed to support and mitigate the coronavirus pandemic along with minimum loss of economic endeavours, and propose a solution to detect the social distancing among people gathered at any public place.

The word “social distancing” is best practice in the direction of efforts through a variety of means, aiming to minimize or interrupt the transmission of COVID-19. It aims at reducing the physical contact between possibly infected individuals and healthy persons. As per the WHO norms~\cite{hensley2020social} it is prescribed that people should maintain at least 6 feet of distance among each other in order to follow social distancing.

\begin{figure}
    \centering
    \includegraphics[scale=0.25] {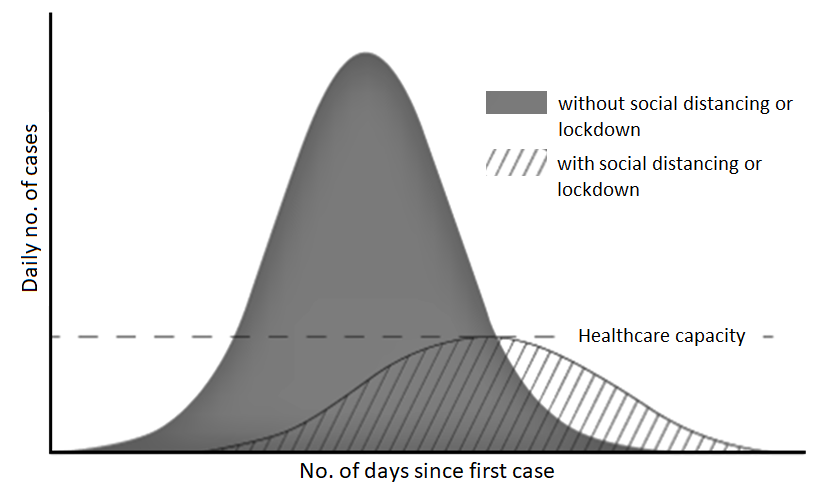}
    \caption{An outcome of social distancing as the reduced peak of the epidemic and matching with available healthcare capacity.}
    \label{fig2}
\end{figure}

A recent study indicates that social distancing is an important containment measure and essential to prevent SARS-CoV-2, because people with mild or no symptoms may fortuitously carry corona infection and can infect others~\cite{sa4}. Fig.~\ref{fig2} indicates that proper social distancing is the best way to reduce infectious physical contact, hence reduces the infection rate~\cite{fong2020nonpharmaceutical, ahmed2018effectiveness}. This reduced peak may surely match with the available healthcare infrastructure and help to offer better facilities to the patients battling against the coronavirus pandemic. 
Epidemiology is the study of factors and reasons for the spread of infectious diseases. To study epidemiological phenomena, mathematical models are always the most preferred choice. Almost all models descend from the classical SIR model of Kermack and McKendrick established in 1927~\cite{kermack1991contributions}. Various research works have been done on the SIR model and its extensions by the deterministic system~\cite{eksin2019systematic}, and consequently, many researchers studied stochastic biological systems and epidemic models~\cite{zhao2016asymptotic}.

Respiratory diseases are infectious where the rate and mode of transmission of the causing virus are the most critical factors to be considered for the treatment or ways to stop the spread of the virus in the community. Several medicine organizations and pandemic researchers are trying to develop vaccines for COVID-19, but still, there is no well-known medicine available for treatment. Hence, precautionary steps are taken by the whole world to restrict the spread of infection. Recently, Eksin et al.~\cite{eksin2019systematic} proposed a modified SIR model with the inclusion of a social distancing parameter, $a(I, R)$ which can be determined with the help of the number of infected and recovered persons represented as $I$ and $R$, respectively.
\begin{equation}
    \begin{aligned}
    \frac{dS}{dt} &= -\beta S \frac{I}{N}a(I,N) \\ \frac{dI}{dt} &= -\delta I+\beta I \frac{I}{N}a(I,N) \\
    \frac{dR}{dt} &= \delta I
    \end{aligned}
    \label{eq1}
\end{equation}
where $\beta$ represents the infection rate and $\delta$ represents recovery rate. The population size is computed as $N = S + I + R$. Here the social distancing term ($a(I,R):{\mathbb{R}}^2 \ \epsilon \ [0,1]$) maps the transition rate from a susceptible state ($S$) to an infected state ($I$), which is calculated by $\frac{a\beta SI}{N}$.

The social distancing models are of two types, where the first model is known as \enquote{long-term awareness} in which the occurrence of interaction of an individual with other is reduced proportionally with the cumulative percentage of affected (infectious and recovered) individuals (Eq.~\ref{eq2}),
\begin{equation}
 a = {\left (1- \frac{I+R}{N} \right)}^k
  \label{eq2}
\end{equation}
Meanwhile, the second model is known as \enquote{short-term awareness}, where the reduction in interaction is directly proportional to the proportion of infectious individuals at a given instance (Eq.~\ref{eq3}),
\begin{equation}
  a = {\left (1- \frac{I}{N} \right)}^k
  \label{eq3}
\end{equation}
where $k$ is behavior parameter defined as, $k \geq 0$. Higher value of $k$ implies that individuals are becoming sensitive to the disease prevalence.

In the similar background, on April 16, 2020, a company Landing AI~\cite{sA21} under the leadership of most recognizable names in AI, Dr. Andrew Ng~\cite{sA22} announced the creation of an AI tool to monitor social distancing at the workplace. In a brief article, the company claimed that the upcoming tool could detect if people are maintaining the safe physical distance from each other by analyzing real-time video streams from the camera. It is also claimed that this tool can easily get integrated with existing security cameras available at different workplaces to maintain a safe distance among all workers. A brief demo was released that shows three steps: calibration, detection and measurement to monitor the social distancing. On April 21, 2020,  Gartner, Inc. identified Landing AI as Cool Vendors in AI Core Technologies to appreciate their timely initiative in this revolutionary area to support the fight against the COVID -19~\cite{sA29}.

Motivated by this, in this present work authors are attempting to check and compare the performance of popular object detection and tracking schemes in monitoring the social distancing. Rest of the paper structure is organized as follows: Section II presents the recent work proposed in this field of study, followed by the state-of-the-art object detection and tracking models in Section III. Later, in Section IV the deep learning based framework is proposed to monitor social distancing. In Section V experimentation and the corresponding results are discussed, accompanied by the outcome in Section VI. In Section VII the future scope and challenges are discussed and lastly Section VIII presents the conclusion of the present research work.

\section{Background study and related work }
Social distancing is surely the most trustworthy technique to stop the spreading of infectious disease, with this belief, in the background of December 2019, when COVID-19 emerged in Wuhan, China, it was opted as an unprecedented measure on January 23, 2020~\cite{sA8}. Within one month, the outbreak in China gained a peak in the first week of February with 2,000 to 4,000 new confirmed cases per day. Later, for the first time after this outbreak, there have been a sign of relief with no new confirmed cases for five consecutive days up to 23 March 2020~\cite{sA10}. This is evident that social distancing measures enacted in China initially, adopted worldwide later to control COVID-19.

Prem et al.~\cite{prem2020effect} aimed to study the effects of social distancing measures on the spread of the COVID-19 epidemic. Authors used synthetic location-specific contact patterns to simulate the ongoing trajectory of the outbreak using susceptible-exposed-infected-removed (SEIR) models. It was also suggested that premature and sudden lifting of social distancing could lead to an earlier secondary peak, which could be flattened by relaxing the interventions gradually~\cite{prem2020effect}. As we all understand, social distancing though essential but economically painful measures to flatten the infection curve. Adolph et al.~\cite{adolph2020pandemic} highlighted the situation of the United States of America, where due to lack of common consent among all policymakers it could not be adopted at an early stage, which is resulting into on-going harm to public health. Although social distancing impacted economic productivity, many researchers are trying hard to overcome the loss. Following from this context, Kylie et al.~\cite{ainslie2020evidence} studied the correlation between the strictness of social distancing and the economic status of the region. The study indicated that intermediate levels of activities could be permitted while avoiding a massive outbreak. 

Since the novel coronavirus pandemic began, many countries have been taking the help of technology based solutions in different capacities to contain the outbreak~\cite{sonbhadra2020target, punn2020automated, Punn2020.04.08.20057679}. Many developed countries, including India and South Korea, for instance, utilising GPS to track the movements of the suspected or infected persons to monitor any possibility of their exposure among healthy people. In India, the government is using the Arogya Setu App, which worked with the help of GPS and bluetooth to locate the presence of COVID-19 patients in the vicinity area. It also helps others to keep a safe distance from the infected person~\cite{sA17}. On the other hand, some law enforcement departments have been using drones and other surveillance cameras to detect mass gatherings of people, and taking regulatory actions to disperse the crowd~\cite{robakowska2017use, 8844927}. Such manual intervention in these critical situations might help flatten the curve, but it also brings a unique set of threats to the public and is challenging to the workforce.

Human detection using visual surveillance system is an established area of research which is relying upon manual methods of identifying unusual activities, however, it has limited capabilities~\cite{sulman2008effective}. In this direction, recent advancements advocate the need for intelligent systems to detect and capture human activities. Although human detection is an ambitious goal, due to a variety of constraints such as low-resolution video, varying articulated pose, clothing, lighting and background complexities and limited machine vision capabilities, wherein prior knowledge on these challenges can improve the detection performance~\cite{wang2013intelligent}. 

Detecting an object which is in motion, incorporates two stages: object detection~\cite{joshi2012survey} and object classification~\cite{javed2002tracking}. The primary stage of object detection could be achieved by using background subtraction~\cite{brutzer2011evaluation}, optical flow~\cite{aslani2013optical} and spatio-temporal filtering techniques~\cite{dollar2005behavior}. In the background subtraction method~\cite{piccardi2004background}, the difference between the current frame and a background frame (first frame), at pixel or block level is computed. Adaptive Gaussian mixture, temporal differencing, hierarchical background models, warping background and non-parametric background are the most popular approaches of background subtraction~\cite{xu2016background}. In optical flow-based object detection technique~\cite{aslani2013optical}, flow vectors associated with the object’s motion are characterised over a time span in order to identify regions in motion for a given sequence of images~\cite{tsutsui2001optical}. Researchers reported that optical flow based techniques consist of computational overheads and are sensitive to various motion related outliers such as noise, colour and lighting, etc.~\cite{agarwal2016review}. In another method of motion detection Aslani et al.~\cite{dollar2005behavior} proposed spatio-temporal filter based approach in which the motion parameters are identified by using three-dimensional (3D) spatio-temporal features of the person in motion in the image sequence. These methods are advantageous due to its simplicity and less computational complexity, however shows limited performance because of noise and uncertainties on moving patterns~\cite{niyogi1994analyzing}. 

Object detection problems have been efficiently addressed by recently developed advanced techniques. In the last decade, convolutional neural networks (CNN), region-based CNN~\cite{zhao2019object} and faster region-based CNN~\cite{krizhevsky2012imagenet} used region proposal techniques to generate the objectness score prior to its classification and later generates the bounding boxes around the object of interest for visualization and other statistical analysis~\cite{ren2015faster}. Although these methods are efficient but suffer in terms of larger training time requirements~\cite{chen2017implementation}. Since all these CNN based approaches utilize classification, another approach YOLO considers a regression based method to dimensionally separate the bounding boxes and interpret their class probabilities~\cite{redmon2016you}. In this method, the designed framework efficiently divides the image into several portions representing bounding boxes along with the class probability scores for each portion to consider as an object. This approach offers excellent improvements in terms of speed while trading the gained speed with the efficiency. The detector module exhibits powerful generalization capabilities of representing an entire image~\cite{ putra2018convolutional}.

Based on the above concepts, many research findings have been reported in the last few years. Crowd counting emerged as a promising area of research, with many societal applications. Eshel et al.~\cite{eshel2008homography}, focused on crowd detection and person count by proposing multiple height homographies for head top detection and solved the occlusions problem associated with video surveillance related applications. Chen et al.~\cite{chen2009online} developed an electronic advertising application based on the concept of crowd counting. In similar application, Chih-Wen et al.~\cite{su2009vision} proposed a vision-based people counting model.  Following this, Yao et al.~\cite{yao2011fast} generated inputs from stationary cameras to perform background subtraction to train the model for the appearance and the foreground shape of the crowd in videos.

Once an object is detected, classification techniques can be applied to identify a human on the basis of shape, texture or motion-based features. In shape-based methods, the shape related  information of moving regions such as points, boxes and blobs are determined to identify the human. This method performs poorly due to certain limitations in standard template-matching schemes~\cite{wu2007detection, eishita2012occlusion}, which is further enhanced by applying part-based template matching~\cite{singh2008human} approach. In another research, Dalal et al.~\cite{dalal2005histograms} proposed texture-based schemes such as histograms of oriented gradient (HOG), which utilises high dimensional features based on edges along with the support vector machine (SVM) to detect humans.

According to recent research, further identification of a person through video surveillance can be done by using face~\cite{huang2010shape, samal1992automatic} and  gait recognition~\cite{cunado1997using} techniques.  However, detection and tracking of people under crowd are difficult sometimes due to partial or full occlusion problems. Leibe et al.~\cite{leibe2005pedestrian} proposed trajectory estimation based solution while Andriluka et al.~\cite{andriluka2008people} proposed a solution to detect partially occluded people using tracklet-based detectors. Many other tracking techniques, including a variety of object and motion representations, are reviewed by Yilmaz et al.~\cite{yilmaz2006object}.

A large number of studies are available in the area of video surveillance. Among many publically available datasets, KTH human motion dataset~\cite{schuldt2004recognizing} shows six categories of activities, whereas INRIA XMAS multi-view dataset~\cite{weinland2006free} and Weizmann human action dataset~\cite{blank2005actions} contain 11 and 10 categories of actions, respectively. Another dataset named as performance evaluation of tracking and surveillance (PETS) is proposed by a group of researchers at university of Oxford~\cite{parkhi2012oxford}. This dataset is available for vision based research comprising a large number of datasets for varying tasks in the field of computer vision. In the present research, in order to fine-tune the object detection and tracking models for identifying the person, open images datasets~\cite{kuznetsova2020open} are considered. It is a collection of 19,957 classes out of which the models are trained for the identification of a person. The images are annotated with image-level labels and corresponding coordinates of the bounding boxes representing the person. Furthermore, the fine tuned proposed framework is simulated on the Oxford town center surveillance footage~\cite{8844927} to monitor social distancing. 

We believe that having a single dataset with unified annotations for image classification, object detection, visual relationship detection, instance segmentation, and multimodal image descriptions will enable us to study and perform object detection tasks efficiently and stimulate progress towards genuine understanding of the scene. All explored literature and related research work clearly establishes a picture that the application of human detection can easily get extended to many applications to cater the situation that arises presently such as to check prescribed standards for hygiene, social distancing, work practices, etc. 

\section{Object detection and tracking models}
\begin{figure}
    \centering
    \includegraphics[scale=0.24] {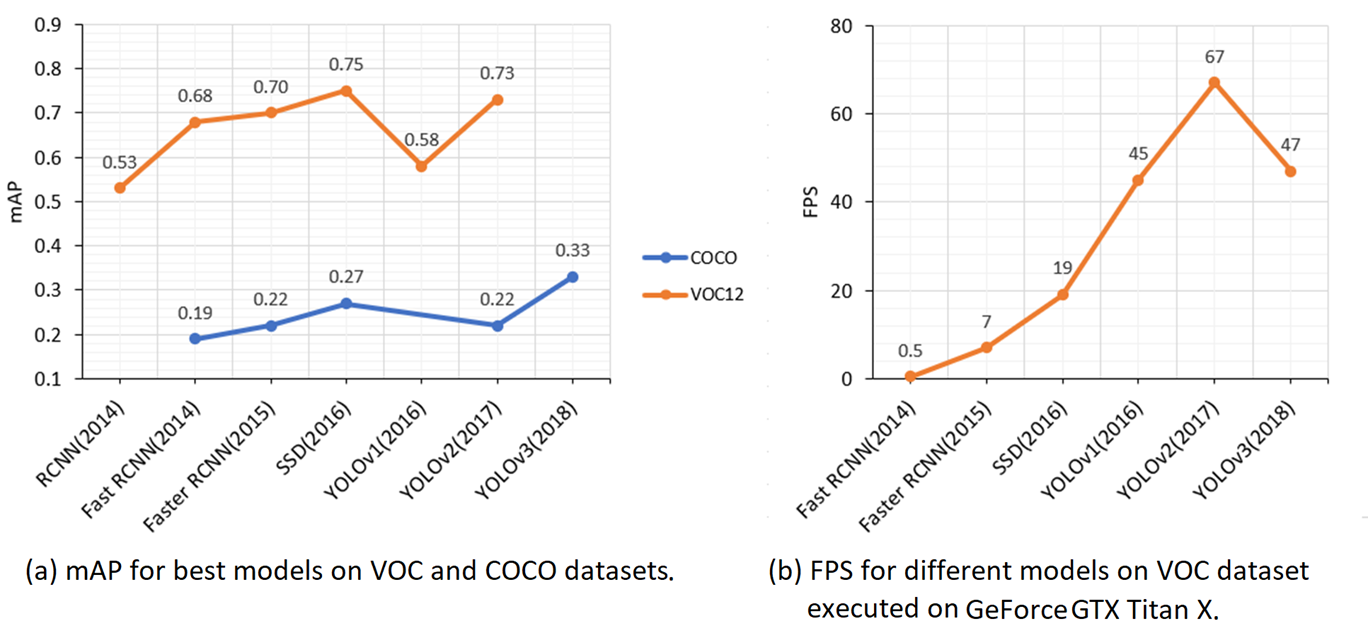}
    \caption{Performance overview of the most popular object detection models on PASCAL-VOC and MS-COCO datasets.
}
    \label{fig3}
\end{figure}

As observed from Fig.~\ref{fig3}, the successful object detection models like RCNN~\cite{girshick2014rich}, fast RCNN~\cite{girshick2015fast}, faster RCNN~\cite{ren2015faster}, SSD~\cite{liu2016ssd}, YOLO v1~\cite{redmon2016you}, YOLO v2~\cite{redmon2017yolo9000} and YOLO v3~\cite{redmon2018yolov3} tested on PASCAL-VOC~\cite{everingham2010pascal} and MS-COCO~\cite{lin2014microsoft} datasets, undergo trade-off between speed and accuracy of the detection which is dependent on various factors like backbone architecture (feature extraction network e.g. VGG-16~\cite{simonyan2014very}, ResNet-101~\cite{he2016deep}, Inception v2~\cite{szegedy2016rethinking}, etc.), input sizes, model depth, varying software and hardware environment. A feature extractor tends to encode the model’s input into certain feature representation which aids in learning and discovering the patterns associated with the desired objects. In order to identify multiple objects of varying scale or size, it also uses predefined boxes covering an entire image termed as anchor boxes. Table~\ref{tab1} describes the performance in terms of accuracy for each of these popular and powerful feature extraction networks on ILSVRC ImageNet challenge~\cite{russakovsky2015imagenet}, along with the number of trainable parameters, which have a direct impact on the training speed and time. As highlighted in Table~\ref{tab1}, the ratio of accuracy to the number of parameters is highest for Inception v2 model indicating that Inception v2 achieved adequate classification accuracy with minimal trainable parameters in contrast to other models, and hence is utilized as a backbone architecture for faster and efficient computations in the faster RCNN and SSD object detection models, whereas YOLO v3 uses different architecture Darknet-53 as proposed by Redmon et al.~\cite{redmon2018yolov3}.

\begin{table}[h!]
    \centering
    \caption{ Performance of the feature extraction network on ImageNet challenge.}
    \begin{tabular}{|p{2.15cm}|c|c|c|}
\hline
Backbone model  & Accuracy (a) & Parameters (p) & Ratio (a*100/p) \\ \hline
VGG-16~\cite{simonyan2014very}  & 0.71                                      & 15 M                                            & 4.73                                           \\ \hline
ResNet-101~\cite{he2016deep} & 0.76                                  & 42.5 M                                          & 1.78                                            \\ \hline
\textbf{Inception v2}~\cite{szegedy2016rethinking}  & \textbf{0.74}                                          & \textbf{10 M}                                            & \textbf{7.40}                                             \\ \hline
Inception v3~\cite{szegedy2017inception} & 0.78 & 22 M                                                     & 3.58                                                      \\ \hline
 Resnet v2~\cite{szegedy2017inception} & 0.80                                                   & 54 M                                                     & 1.48                                                      \\ \hline
\end{tabular}
    \label{tab1}
\end{table}

\subsection{Anchor boxes}
With the exhaustive literature survey, it is observed that every popular object detection model utilizes the concept of anchor boxes to detect multiple objects in the scene~\cite{zhao2019object}. These boxes are overlaid on the input image over various spatial locations (per filter) with varying sizes and aspect ratio. In this article for an image of dimension breadth ($b$) $\times$ height ($h$) the anchor boxes are generated in the following manner. Consider the parameters, size as $p \ \epsilon \ (0,1]$ and aspect ratio as $r > 0$, then the anchor boxes for a certain location in an image can be constructed with dimensions as $bp\sqrt{r} \times hp\sqrt{r}$. Table~\ref{tab2} shows the values of $p$ and $r$ configured for each model. Later the object detection model is trained to predict for each generated anchor box to belong to a certain class, and an offset to adjust the dimensions of the anchor box to better fit the ground-truth of the object while using the classification and regression loss. Since there are many anchor boxes for a spatial location, the object can get associated with more than one anchor box. This problem is dealt with non-max suppression (NMS) by computing intersection over union (IoU) parameter that limits the anchor boxes association with the object of interest by calculating the score as the ratio of overlapping regions between the assigned anchor box and the ground-truth to the union of regions of the anchor box and the ground-truth. The score value is then compared with the set threshold hyperparameter to return the best bounding box for an object.
\begin{table}[]
    \centering
    \caption{Hyperparameters for generating the anchor boxes.}
    \label{tab2}
   \begin{tabular}{|l|l|l|l|l|}
\hline
\begin{tabular}[c]{@{}l@{}}Detection\\ model\end{tabular} & \begin{tabular}[c]{@{}l@{}}Size vector\\ (p)\end{tabular} & \begin{tabular}[c]{@{}l@{}}Aspect ratio\\ (r)\end{tabular} & \begin{tabular}[c]{@{}l@{}}Anchor\\ boxes\end{tabular} & \begin{tabular}[c]{@{}l@{}}IoU th.\\ for NMS\end{tabular} \\ \hline
\begin{tabular}[c]{@{}l@{}}Faster\\ RCNN\end{tabular}     & {[}0.25, 0.5, 1.0{]}                                      & {[}0.5, 1.0, 2.0{]}                                        & 9                                                      & 0.7                                                       \\ \hline
SSD                                                       & {[}0.2, 0.57, 0.95{]}                                     & {[}0.3, 0.5, 1.0{]}                                        & 9                                                      & 0.6                                                       \\ \hline
YOLO v3                                                   & {[}0.25, 0.5, 1.0{]}                                      & {[}0.5, 1.0, 2.0{]}                                        & 9                                                      & 0.7                                                       \\ \hline
\end{tabular}
\end{table}
\subsubsection{Loss Function}
With each step of model training, predicted anchor box \enquote*{a} is assigned a label as positive (1) or negative (0), based on its associativity with the object of interest having ground-truth box \enquote*{g}. The positive anchor box is then assigned a class label $y_o \ \epsilon \  \{c_1, c_2,...., c_n\}$, here $c_n$ indicates the category of the $n^{th}$ object, while also generating the encoding vector for box \enquote*{g} with respect to \enquote*{a} as $f(g_a|a)$, where $y_o = 0$ for negative anchor boxes. Consider an image $I$, for some anchor \enquote*{a}, model with trained parameters $\omega$, predicted the object class as $Y_{cls}(I|a;\omega)$ and the corresponding box offset as $Y_{reg}(I|a;\omega)$,  then the loss for a single anchor prediction can be computed ($L_{cls}$) and bounding box regression loss ($L_{reg}$), as given by the Eq~\ref{eq4}.
\begin{equation}
    \begin{aligned}
    L(a|I; \omega) =\alpha. 1_{a}^{obj} L_{reg}(f(g_a|a) - Y_{reg}(I|a;\omega)) + \\ \beta. L_{cls} (y_a, Y_{cls}(I|a;\omega))
    \end{aligned}
    \label{eq4}
\end{equation}
where $1_{a}^{obj}$ is 1 if \enquote*{a} is a positive anchor, $\alpha$ and $\beta$ are the weights associated with the regression and classification loss. Later, the overall loss of the model can be computed as the average of the $L(a|I;w)$ over the predictions for all the anchors. 
\subsection{Faster RCNN}
\begin{figure}
    \centering
    \includegraphics[scale=0.125] {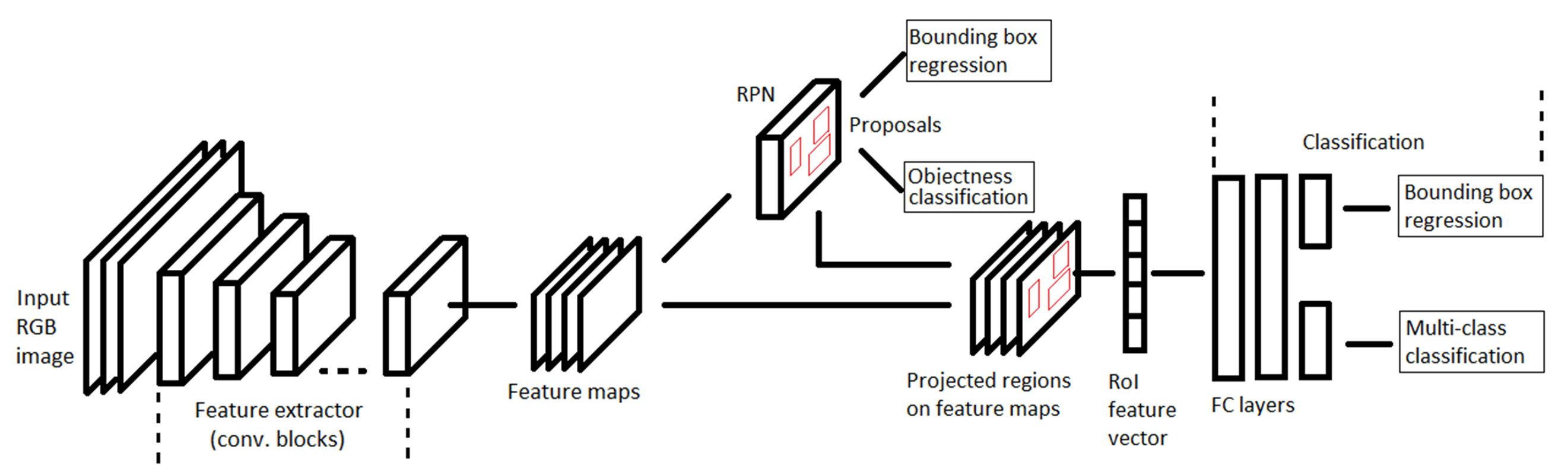}
    \caption{Schematic representation of faster RCNN architecture}
    \label{fig4}
\end{figure}

Proposed by Ren et al.~\cite{ren2015faster}, the faster RCNN is derived from its predecessors RCNN~\cite{girshick2014rich} and fast RCNN~\cite{girshick2015fast}, which rely on external region proposal approach based on selective search (SS)~\cite{google}. Many researchers~\cite{punn2020inception, vaswani2018tensor2tensor, amodei2016deep}, observed that instead of using the SS, it is recommended to utilize the advantages of convolution layers for better and faster localization of the objects. Hence, Ren et al. proposed the Region Proposal Network (RPN) which uses CNN models, e.g. VGGNet, ResNet, etc. to generate the region proposals that made faster RCNN 10 times faster than fast RCNN. Fig.~\ref{fig4} shows the schematic representation of faster RCNN architecture, where RPN module performs binary classification of an object or not an object (background) while classification module assigns categories for each detected object (multi-class classification) by using the region of interest (RoI) pooling~\cite{ren2015faster} on the extracted feature maps with projected regions.
\subsubsection{Loss function}
The faster RCNN is the combination of two modules RPN and fast RCNN detector. The overall multi-task loss function is composed of classification loss and bounding box regression loss as defined in Eq.~\ref{eq4} with $L_{cls}$ and $L_{reg}$ functions defined in Eq.~\ref{eq5}
\begin{equation}
    \begin{aligned}
    L_{cls} (p_i, p_{i}^{*}) &= -p_{i}^{*} \log (p_i) - (1-p_{i}^{*}) \log (1- p_i) \\
    L_{reg} (t^u, v) & = \sum_{x \epsilon {x, y, w, h} } L_{1}^{smooth}(t_{i}^{u} -v)\\
     L_{1}^{smooth}(q)&=\begin{cases}
    0.5 q^2, & if \mid q \mid < 1 .\\
    \mid q \mid - 0.5, & \text{otherwise}.
  \end{cases}		
    \end{aligned}
    \label{eq5}
\end{equation}
where $t^u$ is the predicted corrections of the bounding box $t^u = \{t_{x}^{u}, t_{y}^{u}, t_{w}^{u}, t_{h}^{u}\}$. Here $u$ is a true class label, ($x$, $y$) corresponds to the top-left coordinates of the bounding box with height $h$ and width $w$, $v$ is a ground-truth bounding box, $p_{i}^{*}$ is the predicted class and $p_i$ is the actual class, 
\subsection{Single Shot Detector (SSD)}
\begin{figure}
    \centering
    \includegraphics[scale=0.1288]{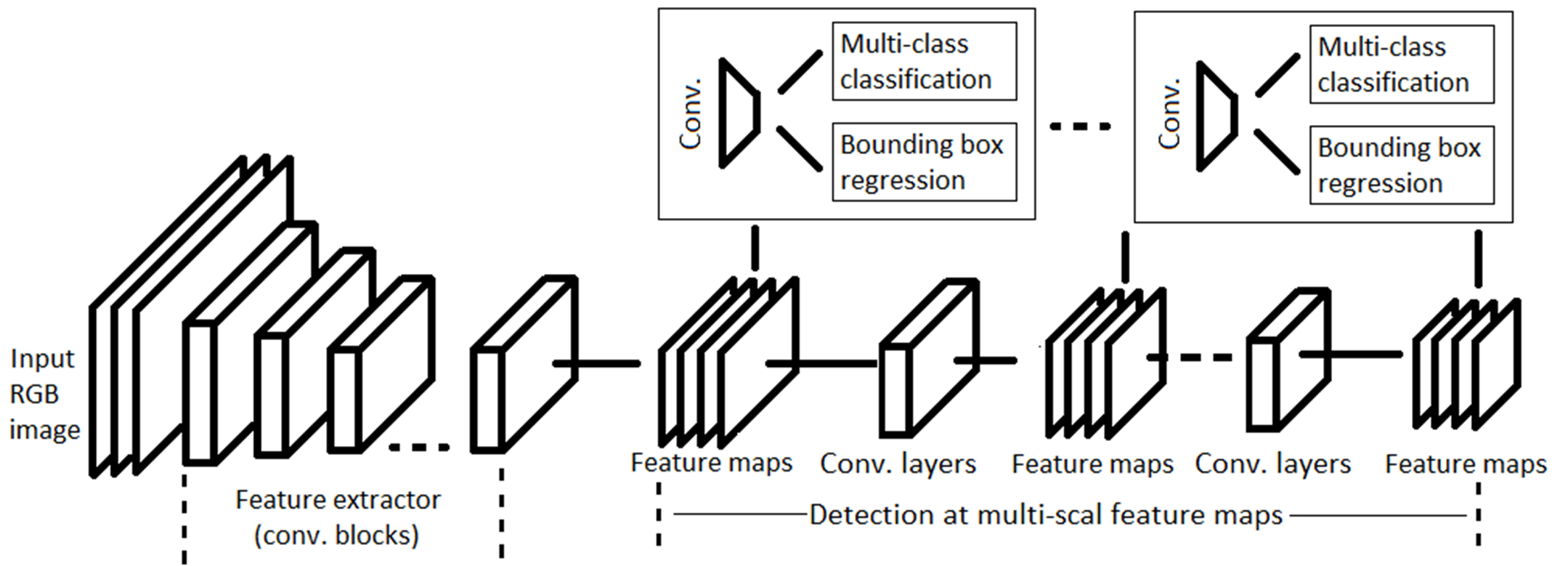}
    \caption{Schematic representation of SSD architecture}
    \label{fig5}
\end{figure}

In this research, single shot detector (SSD)~\cite{liu2016ssd} is also used as another object identification method to detect people in real-time video surveillance system. As discussed earlier, faster R-CNN works on region proposals to create boundary boxes to indicate objects, shows better accuracy, but has slow processing of frames per second (FPS). For real-time processing, SSD further improves the accuracy and FPS by using multi-scale features and default boxes in a single process. It follows the principle of the feed-forward convolution network which generates bounding boxes of fixed sizes along with a score based on the presence of object class instances in those boxes, followed by NMS step to produce the final detections. Thus, it consists of two steps: extracting feature maps and applying convolution filters to detect objects by using an architecture having three main parts. First part is a base pretrained network to extract feature maps, whereas, in the second part, multi-scale feature layers are used in which series of convolution filters are cascaded after the base network. The last part is a non-maximum suppression unit for eliminating overlapping boxes and one object only per box. The architecture of SSD is shown in Fig.~\ref{fig5}.
\subsubsection{Loss function}
Similar to the above discussed faster RCNN model, the overall loss function of the SSD model is equal to the sum of multi-class classification loss ($L_{cls}$) and bounding box regression loss (localization loss, $L_{reg}$), as shown in Eq.~\ref{eq4}, where $L_{reg}$ and $L_{cls}$ is defined by Eq.~\ref{eq6} and~\ref{eq7}:
\begin{equation}
    \begin{aligned}
        L_{reg}(x,l,g) &=\sum_{i \epsilon pos}^{N} \sum_{m \epsilon c_x, c_y, w, h} x_{ij}^{k} {smooth}_{L_1}(l_{i}^{m} -{\hat{g}}_{j}^{m}) ,\\
         {\hat{g}}_{j}^{c_x} &= \frac{ (g_{j}^{c_x}-a_{i}^{c_x})}{a_i^w} , \ 
         {\hat{g}}_{j}^{c_y} = \frac{ (g_{j}^{c_y}-a_{i}^{c_y})}{a_i^h} ,\\
         {\hat{g}}_{j}^{w} &= \log \left ( \frac{g_{j}^{w}}{a_{i}^{w}} \right ) ,\ 
         {\hat{g}}_{j}^{h} = \log \left ( \frac{g_{j}^{h}}{a_{i}^{h}} \right ),\\
          x_{ij}^{p} & =\begin{cases}
    1, & \text{if IoU} > {0.5}\\
    0, & \text{otherwise}.
  \end{cases}	
    \end{aligned}
    \label{eq6}
\end{equation}
where $l$ is the predicted box, $g$ is the ground truth box, $x_{ij}^{p}$ is an indicator that matches the $i^{th}$ anchor box to the $j^{th}$ ground truth box, $c_x$ and $c_y$ are offsets to the anchor box $a$.
\begin{equation}
    \begin{aligned}
    L_{cls}(x,c) &= -\sum_{i \epsilon Pos}^{N} x_{ij}^{p} \log ({\hat{c}}_{i}^{p}) -\sum_{i \epsilon Neg} \log ({\hat{c}}_{i}^{o})
    \end{aligned}
    \label{eq7}
\end{equation}
where ${\hat{c}}_{i}^{p} = \frac{\exp{c_{i}^{p}}}{\sum_{p} \exp{c_{i}^{p}}}$ and $N$ is the number of default matched boxes.
\subsection{YOLO}
\begin{figure}
    \centering
    \includegraphics[scale=0.17]{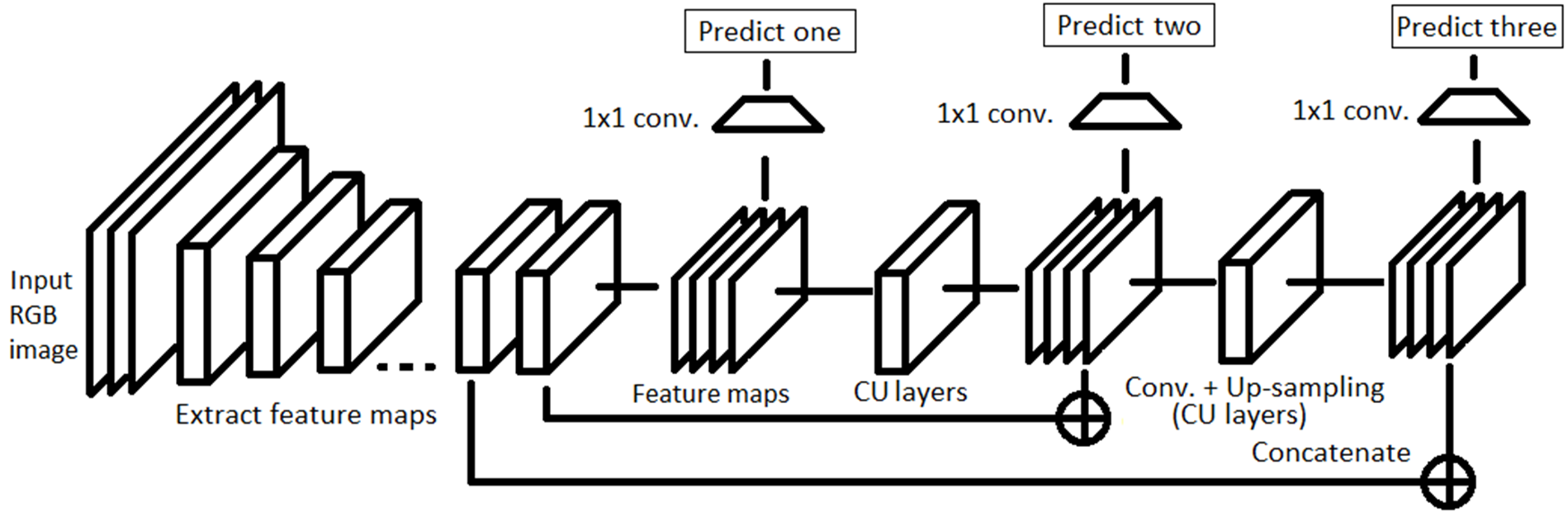}
    \caption{Schematic representation of YOLO v3 architecture}
    \label{fig6}
\end{figure}

For object detection, another competitor of SSD is YOLO~\cite{redmon2016you}. This method can predict the type and location of an object by looking only once at the image. YOLO considers the object detection problem as a regression task instead of classification to assign class probabilities to the anchor boxes. A single convolutional network simultaneously predicts multiple bounding boxes and class probabilities. Majorly, there are three versions of YOLO: v1, v2 and v3. YOLO v1 is inspired by GoogleNet (Inception network) which is designed for object classification in an image. This network consists of 24 convolutional layers and 2 fully connected layers. Instead of the Inception modules used by GoogLeNet, YOLO v1 simply uses a reduction layer followed by convolutional layers. Later, YOLO v2~\cite{redmon2017yolo9000} is proposed with the objective of improving the accuracy significantly while making it faster. YOLO v2 uses Darknet-19 as a backbone network consisting of 19 convolution layers along with 5 max pooling layers and an output softmax layer for object classification. YOLO v2 outperformed its predecessor (YOLO v1) with significant improvements in mAP, FPS and object classification score. In contrast, YOLO v3 performs multi-label classification with the help of logistic classifiers instead of using softmax as in case of YOLO v1 and v2. In YOLO v3 Redmon et al. proposed Darknet-53 as a backbone architecture that extracts features maps for classification. In contrast to Darknet-19, Darknet-53 consists of residual blocks (short connections) along with the upsampling layers for concatenation and added depth to the network. YOLO v3 generates three predictions for each spatial location at different scales in an image, which eliminates the problem of not being able to detect small objects efficiently~\cite{SK4}. Each prediction is monitored by computing objectness, boundary box regressor and classification scores. In Fig.~\ref{fig6} a schematic description of the YOLOv3 architecture is presented.
\subsubsection{Loss function}
The overall loss function of YOLO v3 consists of localization loss (bounding box regressor), cross entropy and confidence loss for classification score, defined as follows:
\begin{equation}
    \begin{aligned}
   {\lambda}_{coord} \sum_{i=0}^{S^2} \sum_{j=0}^{B} {{1}}_{i,j}^{obj} ({(t_x - {\hat{t}}_x)}^2 + {(t_y - {\hat{t}}_y)}^2 +{(t_w - {\hat{t}}_w)}^2 + \\ {(t_h - {\hat{t}}_h)}^2) \\
  + \sum_{i=0}^{S^2} \sum_{j=0}^{B} {{1}}_{i,j}^{obj} (-\log(\sigma(t_o)) + \sum_{k=1}^{C} BCE({\hat{y}}_k, \sigma (s_k)))
\\
+
{\lambda}_{noobj}\sum_{i=0}^{S^2} \sum_{j=0}^{B} {{1}}_{i,j}^{noobj} (-\log(1- \sigma(t_o))
    \end{aligned}
    \label{eq9}
\end{equation}
where ${\lambda}_{coord}$ indicates the weight of the coordinate error, ${S^2}$ indicates the number of grids in the image, and $B$ is the number of generated bounding boxes per grid. ${1}_{i,j}^{obj} = 1$ describes that object confines in the $j^{th}$ bounding box in grid $i$, otherwise it is $0$.
\subsection{Deepsort}
Deepsort is a deep learning based approach to track custom objects in a video~\cite{wojke2017simple}. In the present research,  Deepsort is utilized to track individuals present in the surveillance footage. It makes use of patterns learned via detected objects in the images which is later combined with the temporal information for predicting associated trajectories of the objects of interest. It keeps track of each object under consideration by mapping unique identifiers for further statistical analysis. Deepsort is also useful to handle associated challenges such as occlusion, multiple viewpoints, non-stationary cameras and annotating training data. For effective tracking, the Kalman filter and the Hungarian algorithm are used.  Kalman filter is recursively used for better association, and it can predict future positions based on the current position. Hungarian algorithm is used for association and id attribution that identifies if an object in the current frame is the same as the one in the previous frame.  Initially, a Faster RCNN is trained for person identification and for tracking, a linear constant velocity model~\cite{wojke2018deep} is utilized to describe each target with eight dimensional space as follows: 
\begin{equation}
    x = {[u, v, \lambda, h, x^{,}, y^{,}, {\lambda}^{,} , h^{,} ]}^T
\end{equation}
where ($u,v$) is the centroid of the bounding box, $a$ is the aspect ratio and $h$ is the height of the image. The other variables are the respective velocities of the variables. Later, the standard Kalman filter is used with constant velocity motion and linear observation model, where the bounding coordinates ($u, v, \lambda, h$) are taken as direct observations of the object state.

For each track $k$, starting from the last successful measurement association $a_k$, the total number of frames are calculated. With positive prediction from the Kalman filter, the counter is incremented and later when the track gets associated with a measurement it resets its value to $0$. Furthermore, if the identified tracks exceed a predefined maximum age, then those objects are considered to have left the scene and the corresponding track gets removed from the track set. And if there are no tracks available for some detected objects then new track hypotheses are initiated for each unidentified track of novel detected objects that cannot be mapped to the existing tracks. For the first three frames the new tracks are classified as indefinite until a successful measurement mapping is computed. If the tracks are not successfully mapped with measurement then it gets deleted from the track set. Hungarian algorithm is then utilized in order to solve the mapping problem between the newly arrived measurements and the predicted Kalman states by considering the motion and appearance information with the help of Mahalanobis distance computed between them as defined in Eq.~\ref{eq10}.
\begin{equation}
d^{(1)} (i,j) = {(d_j -y_i)}^T S_{i}^{-1}(d_j -y_i)
    \label{eq10}
\end{equation}
where the projection of the $i^{th}$ track distribution into measurement space is represented by ($y_i ,S_i$) and the $j^{th}$ bounding box detection by $d_j$. The Mahalanobis distance considers this uncertainty by estimating the count of standard deviations, the detection is away from the mean track location. Further, using this metric, it is possible to exclude unlikely associations by thresholding the Mahalanobis distance.  This decision is denoted with an indicator that evaluates to 1 if the association between the $i^{th}$ track and $j^{th}$ detection is admissible (Eq.~\ref{eq11}).
\begin{equation}
b_{i,j}^{(1)} = 1 [d^{(1)} (i,j) < t^{(1)} ]
    \label{eq11}
\end{equation}

Though Mahalanobis distance performs efficiently but fails in the environment where camera motion is possible, thereby another metric is introduced for the assignment problem. This second metric measures the smallest cosine distance between the $i^{th}$ track and $j^{th}$  detection in appearance space as follows:
\begin{equation}
d^{(2)} (i, j)= min\{1- {r_j}^T {r_k}^{(i)} \mid {r_k}^{(i)} \  \epsilon  \ {\mathbb{R}}^2 \}
    \label{eq12}
\end{equation}
Again, a binary variable is introduced to indicate if an association is admissible according to the following metric:
\begin{equation}
b_{i,j}^{(1)} = 1 [d^{(2)} (i,j) < t^{(2)} ]
    \label{eq13}
\end{equation}
and a suitable threshold is measured for this indicator on a separate training dataset. To build the association problem, both metrics are combined using a weighted sum:
\begin{equation}
c_{i,j} = \lambda d^{(1)} (i,j) + (1 - \lambda) d^{(2)} (i, j)
\end{equation}
where an association is admissible if it is within the gating region of both metrics: 
\begin{equation}
b_{i,j} = \prod_{m=1}{2} b_{i,j}^{(m)}.
\end{equation}
The influence of each metric on the combined association cost can be controlled through hyperparameter $\lambda$. 

\section{Proposed approach}
The emergence of deep learning has brought the best performing techniques for a wide variety of tasks and challenges including medical diagnosis~\cite{punn2020inception}, machine translation~\cite{vaswani2018tensor2tensor}, speech recognition~\cite{amodei2016deep}, and a lot more~\cite{pouyanfar2018survey}. Most of these tasks are centred around object classification, detection, segmentation, tracking, and recognition~\cite{brunetti2018computer, punn2019crowd}. In recent years, the convolution neural network (CNN) based architectures have shown significant performance improvements that are leading towards the high quality of object detection, as shown in Fig.~\ref{fig3}, which presents the performance of such models in terms of mAP and FPS on standard benchmark datasets, PASCAL-VOC~\cite{everingham2010pascal} and MS-COCO~\cite{lin2014microsoft}, and similar hardware resources.

In the present article, a deep learning based framework is proposed that utilizes object detection and tracking models to aid in the social distancing remedy for dealing with the escalation of COVID-19 cases. In order to maintain the balance of speed and accuracy, YOLO v3~\cite{redmon2018yolov3} alongside the Deepsort~\cite{wojke2017simple} are utilized as object detection and tracking approaches while surrounding each detected object with the bounding boxes. Later, these bounding boxes are utilized to compute the pairwise \textit{L2} norm with computationally efficient vectorized representation for identifying the clusters of people not obeying the order of social distancing. Furthermore, to visualize the clusters in the live stream, each bounding box is color-coded based on its association with the group where people belonging to the same group are represented with the same color. Each surveillance frame is also accompanied with the streamline plot depicting the statistical count of the number of social groups and an index term (violation index) representing the ratio of the number of people to the number of groups. Furthermore, estimated violations can be computed by multiplying the violation index with the total number of social groups.
\subsection{ Workflow}
This section includes the necessary steps undertaken to compose a framework for monitoring social distancing.
\begin{itemize}
    \item[1.] Fine-tune the trained object detection model to identify and track the person in a footage.
    \item[2.] The trained model is feeded with the surveillance footage. The model generates a set of bounding boxes and an ID for each identified person.
    \item [3.] Each individual is associated with three-dimensional feature space ($x, y, d$), where ($x$, $y$)  corresponds to the centroid coordinates of the bounding box and $d$ defines the depth of the individual as observed from the camera.
    \begin{equation}
    d = ((2 * 3.14 * 180) / (w + h * 360) * 1000 + 3)
    \end{equation}
    where $w$ is the width of the bounding box and $h$ is the height of the bounding box~\cite{kaggle}.
    \item[4.] For the set of bounding boxes, pairwise \textit{L2} norm is computed as given by the following equation.
    \begin{equation}
        ||D||_2=\sqrt{\sum_{i=1}^{n} {(q_i -p_i)}^2}
    \end{equation}
    where in this work $n = 3$.
    \item[5.] The dense matrix of \textit{L2} norm is then utilized to assign the neighbors for each individual that satisfies the closeness sensitivity. With extensive trials the closeness threshold is updated dynamically based on the spatial location of the person in a given frame ranging between ($90, 170$) pixels.
    \item[6.] Any individual that meets the closeness property is assigned a neighbour or neighbours forming a group represented in a different color coding in contrast to other people.
    \item[7.] The formation of groups indicates the violation of the practice of social distancing which is quantified with help of the following:
     \begin{itemize}
      \item Consider $n_g$ as number of groups or clusters identified, and $n_p$ as total number of people found in close proximity.
      \item $v_i = n_p/n_g$, where $v_i$ is the violation index.
      \end{itemize}
\end{itemize}
\section{Experiments and results}
\begin{figure}
    \centering
    \includegraphics[scale=0.24] {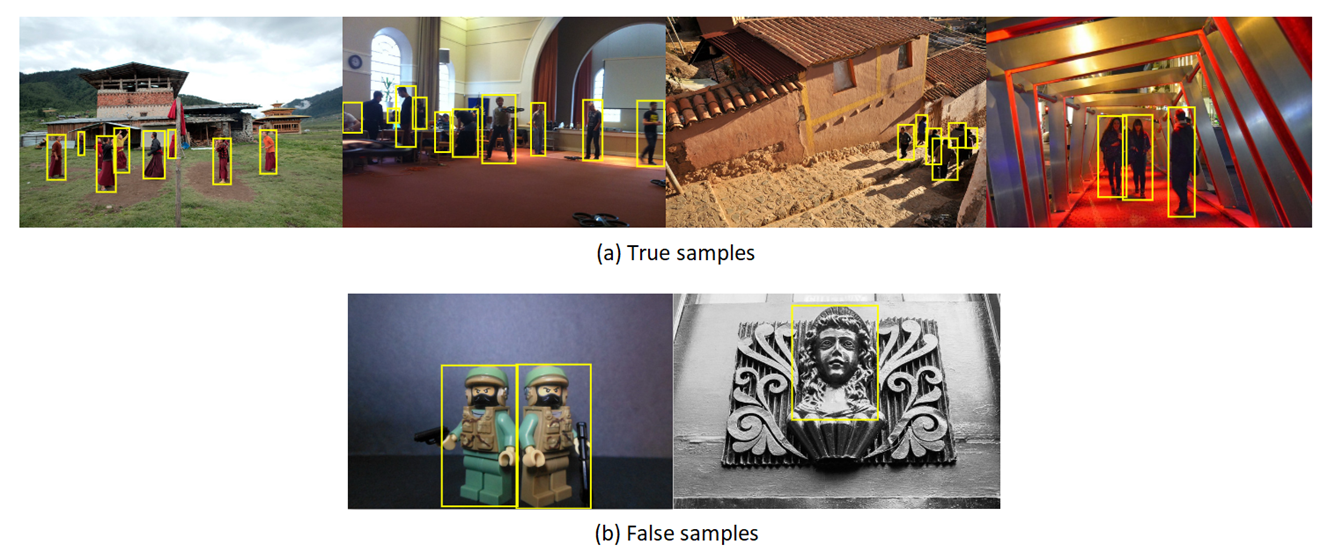}
    \caption{Data samples showing (a) true samples and (b) false samples of a \enquote{Person} class from the open image dataset.
}
    \label{fig7}
\end{figure}

\begin{figure}
    \centering
    \includegraphics[scale=0.17] {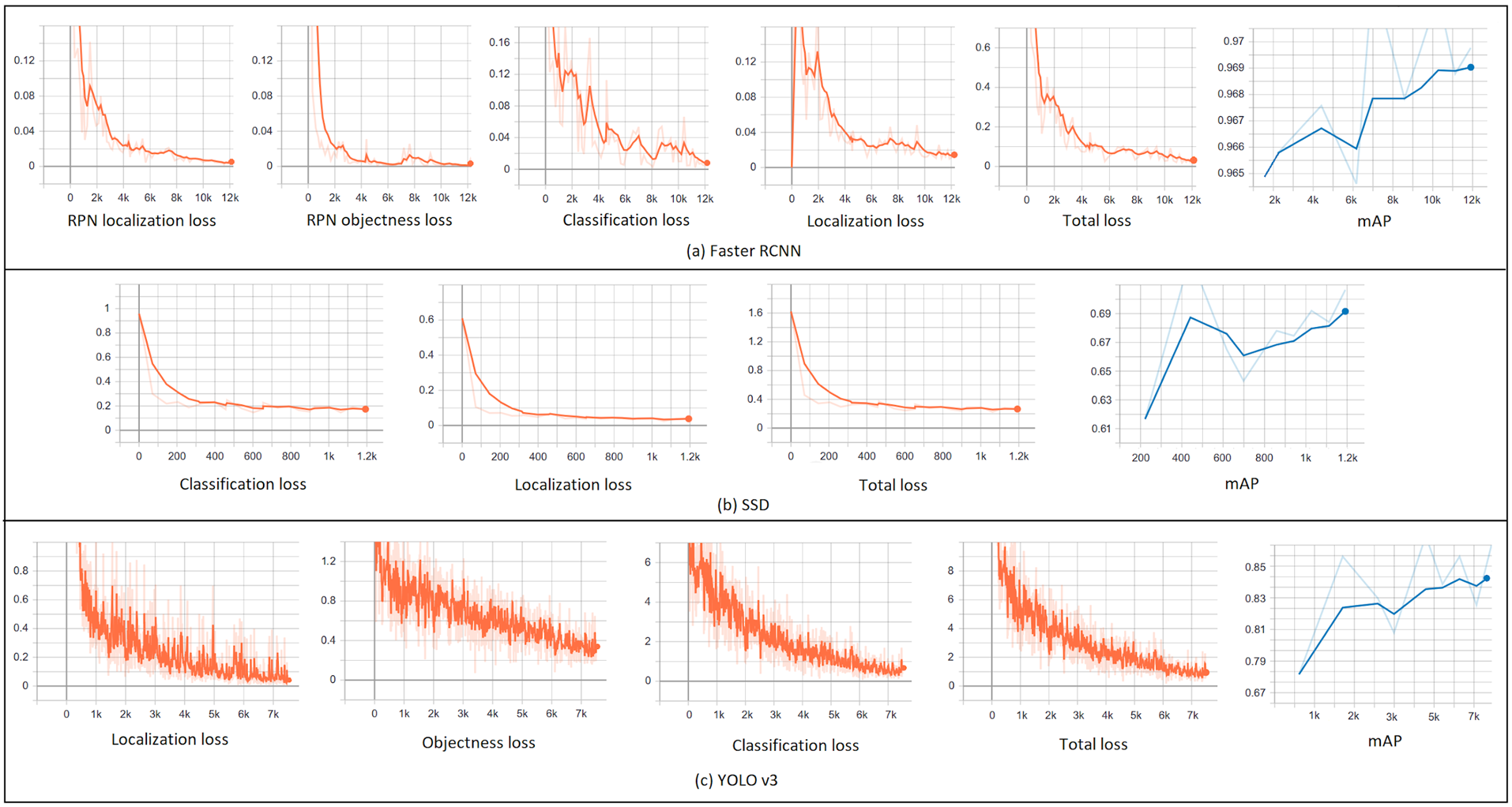}
    \caption{Losses per iteration of the object detection models during the training phase on the OID validation set for detecting the person in an image.
}
    \label{fig8}
\end{figure}

The above discussed object detection models are fine tuned for binary classification (person or not a person) with Inception v2 as a backbone network on the Nvidia GTX 1060 GPU, using the dataset acquired from the open image dataset (OID) repository~\cite{google} maintained by the Google open source community. The diverse images with a class label as “Person” are downloaded via OIDv4 toolkit~\cite{megapixels} along with the annotations. Fig.~\ref{fig7} shows the sample images of the obtained dataset consisting of 800 images which is obtained by manually filtering  to only contain the true samples. The dataset is then divided into training and testing sets, in 8:2 ratio. In order to make the testing robust, the testing set is also accompanied by the frames of surveillance footage of the Oxford town center~\cite{8844927}. Later this footage is also utilized to simulate the overall approach for monitoring the social distancing.
In case of faster RCNN, the images are resized to $P$ pixels on the shorter edge with 600 and 1024 for low and high resolution, while in SSD and YOLO the images are scaled to the fixed dimension $P\times P$ with $P$ value as 416. During the training phase, the performance of the models is continuously monitored using the mAP along with the localization, classification and overall loss in the detection of the person as indicated in Fig.~\ref{fig8}. Table~\ref{tab3} summarizes the results of each model obtained at the end of the training phase with the training time (TT), number of iterations (NoI), mAP, and total loss (TL) value. It is observed that the faster RCNN model achieved minimal loss with maximum mAP, however, has the lowest FPS, which makes it not suitable for real-time applications. Furthermore, as compared to SSD, YOLO v3 achieved better results with balanced mAP, training time, and FPS score. The trained YOLO v3 model is then utilized for monitoring the social distancing on the surveillance video.

\section{Output}
\begin{figure}
    \centering
    \includegraphics[scale=0.22] {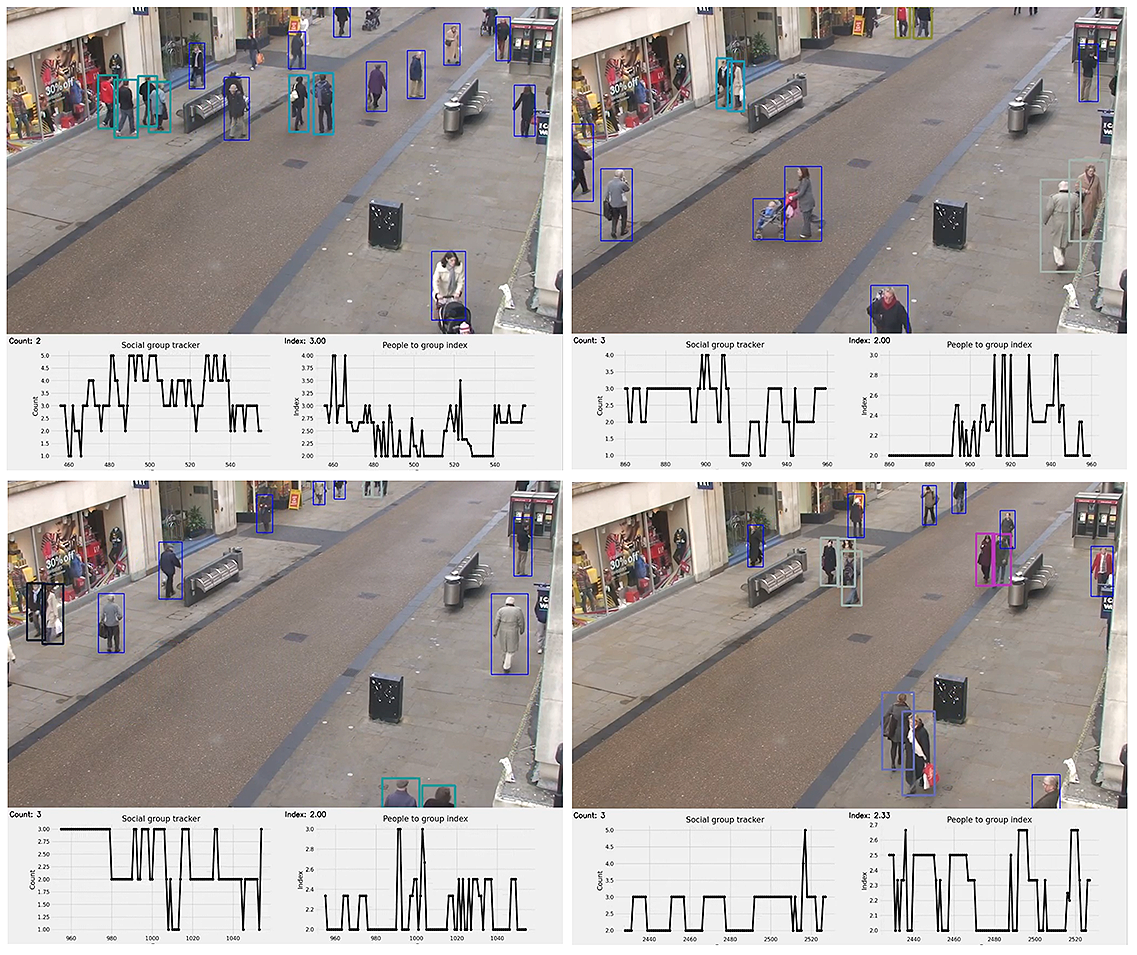}
    \caption{Sample output of the proposed framework for monitoring social distancing on surveillance footage of Oxford Town Center.
}
    \label{fig9}
\end{figure}

The proposed framework outputs (as shown in Fig.~\ref{fig9}) the processed frame with the identified people confined in the bounding boxes while also simulating the statistical analysis showing the total number of social groups displayed by same color encoding and a violation index term computed as the ratio of the number of people to the number of groups. The frames shown in Fig.~\ref{fig9} displays violation index as 3, 2, 2, and 2.33. The frames with detected violations are recorded with the timestamp for future analysis. 
\begin{table}[]
    \centering
    \caption{Performance comparison of the object detection models.}
    \label{tab3}
   \begin{tabular}{|l|l|l|l|l|l|}
\hline
Model            & TT (in sec.)  & NoI           & mAP            & TL            & FPS         \\ \hline
Faster RCNN      & 9651          & 12135         & 0.969          & 0.02          & 3           \\ \hline
SSD              & 2124          & 1200          & 0.691          & 0.22          & 10          \\ \hline
\textbf{YOLO v3} & \textbf{5659} & \textbf{7560} & \textbf{0.846} & \textbf{0.87} & \textbf{23} \\ \hline
\end{tabular}
\end{table}

\section{Future scope and challenges}
Since this application is intended to be used in any working environment; accuracy and precision are highly desired to serve the purpose. Higher number of false positive may raise discomfort and panic situation among people being observed. There may also be genuinely raised concerns about privacy and individual rights which can be addressed with some additional measures such as prior consents for such working environments, hiding a person’s identity in general, and maintaining transparency about its fair uses within limited stakeholders.

\section{Conclusion}
The article proposes an efficient real-time deep learning based framework to automate the process of monitoring the social distancing via object detection and tracking approaches, where each individual is identified in the real-time with the help of bounding boxes. The generated bounding boxes aid in identifying the clusters or groups of people satisfying the closeness property computed with the help of pairwise vectorized approach. The number of violations are confirmed by computing the number of groups formed and violation index term computed as the ratio of the number of people to the number of groups. The extensive trials were conducted with popular state-of-the-art object detection models: Faster RCNN, SSD, and YOLO v3, where YOLO v3 illustrated the efficient performance with balanced FPS and mAP score. Since this approach is highly sensitive to the spatial location of the camera, the same approach can be fine tuned to better adjust with the corresponding field of view.

\section*{Acknowledgment}
The authors gratefully acknowledge the helpful comments and suggestions of colleagues. Authors are also indebted to Interdisciplinary Cyber Physical Systems (ICPS) Programme, Department of Science and Technology (DST), Government of India (GoI) vide Reference No.244 for their financial support to carry out the background research which helped significantly for the implementation of present research work.

\bibliographystyle{IEEEtran}
\bibliography{distance}
\end{document}